\documentclass[letterpaper]{article} 
\usepackage{aaai2027}  
\usepackage[hyphens]{url}  
\usepackage{graphicx} 
\usepackage{natbib}  
\usepackage{caption} 
\usepackage{algorithm}
\usepackage{algorithmic}
\usepackage{amssymb}
\usepackage{amsmath}

\usepackage{array}
\usepackage{caption}
\usepackage{subcaption}

\usepackage{newfloat}
\usepackage{listings}
\DeclareCaptionStyle{ruled}{labelfont=normalfont,labelsep=colon,strut=off} 
\floatstyle{ruled}
\newfloat{listing}{tb}{lst}{}
\floatname{listing}{Listing}

\usepackage{booktabs}

\title{Scalable Kronecker-Fisher Approximation: Efficient Hessian Analysis for Billion-Parameter Language Models Compression}
\author{
    Viacheslav Yusupov\textsuperscript{\rm 1}\corresponding,
    Daria Cherniuk \textsuperscript{\rm 2},
    Evgeny Frolov\textsuperscript{\rm 2, \rm 1}
}
\affiliations{
    \textsuperscript{\rm 1}HSE University\\
    \textsuperscript{\rm 2}AXXX\\
    v.yusupov.lab@gmail.com
}

\begin{document}

\maketitle

\begin{abstract}

In this paper, we propose a scalable Kronecker-based approximation that captures cross-layer interactions without storing the entire Fisher matrix, enabling practical Hessian analysis for billion-parameter networks where full computation is infeasible. Our approach reveals consistent vulnerability patterns: value projection layers exhibit the highest sensitivity and strongest cross-layer correlations across multiple model families, while other components exhibit architecture-specific behaviors. Through extensive experiments on quantization, sparsification, inter-layer corruption, and post-corruption fine-tuning, we demonstrate that our approximation strongly correlates with both performance degradation and recovery. Our framework provides a practical, theoretically grounded tool for identifying fragile components in large models, opening new avenues for guided compression and optimization strategies, such as mixed-precision allocation, layer-wise sparsity, and adaptive low-rank decomposition across layers and even individual weight groups.

\end{abstract}

\section{Introduction}

Understanding the curvature of the loss landscape is fundamental to deep learning, underpinning
principled advances in optimization \citep{izmailov2014newton, zhao2025second}, model compression
\citep{frantar2022gptq}, fine-tuning \citep{liu2024sophia}, and interpretability
\citep{wu2025large}. This curvature is captured by the Hessian matrix and its expectation, the
Fisher information matrix, which indicate which parameters matter most and how perturbations
propagate through the model.

Due to complexity of computing and analyzing Hessian and Fisher matrix for large models, many practical models neglect cross-layer interactions and study and transform layers independently \citep{chekalina2025generalized} or use only diagonal or block diagonal \citep{zhang2017block} part of Hessian for further applications.
Theoretical work that does study the full structure is instead
restricted to small models on synthetic data \citep{dong2025towards} or small datasets (e.g. MNIST \cite{lecun1998mnist}). 
Bridging this gap requires approximations that are both principled and scalable.

We propose such an approximation, based on the Kronecker factorization framework. We show that the Fisher matrix admits an exact decomposition into a sum
of Kronecker products, and that truncating it, together with an exact diagonal and a matrix-free
eigensolver, reduces memory complexity from quadratic to linear in model size while retaining
substantially richer structure than diagonal and block-diagonal methods. Using this approximation, we provide direct
empirical evidence of non-diagonal Hessian structure in LLMs, confirming theoretical predictions
\citep{dong2025towards} at a scale where the full matrix cannot be computed.

Our contributions are as follows:
\begin{itemize}
    \item We propose a scalable Kronecker-based approximation of the Fisher matrix that reduces
    memory complexity from quadratic to linear in model size.
    \item We provide empirical evidence of non-diagonal Hessian structure in large language models,
    validated on four LLMs from 350M to
    7B parameters.
    \item We demonstrate a strong correlation between layer-wise Hessian values and layer
    sensitivity to quantization and sparsification.
    \item We show that off-diagonal curvature predicts inter-layer effects: coupled layers suffer
    excess damage when compressed jointly, beyond the sum of their individual contributions, and fine-tuning the layers most strongly coupled to the corrupted
    ones recovers the most performance.
\end{itemize}

\section{Related Work}
\label{sec:related_work}

\paragraph{Curvature approximation in deep learning.}
Exact second-order information is central to Newton-type optimization
\citep{izmailov2014newton}, and closed-form expressions for network Hessians
are well understood \citep{naumov2017feedforward, botev2017practical}, but the
full Hessian or Fisher information matrix (FIM) is quadratic in the parameter
count and thus intractable beyond small models. Scalable methods therefore
impose structure: Hessian-free optimization avoids materializing the matrix
via Hessian--vector products \citep{martens2010deep}; K-FAC approximates the
per-layer Fisher as a Kronecker product \citep{martens2015optimizing,
van2000ubiquitous}; and block-diagonal schemes discard all inter-layer blocks
\citep{collobert2004large, zhang2017block, dangel2020modular}. At the largest
scales, practice degrades further to diagonal curvature, either implicitly
through adaptive first-order methods \citep{kingma2014adam, das2024towards,
zhang2025adam} or explicitly through diagonal Hessian estimates
\citep{yao2021adahessian, liu2024sophia}, with low-rank preconditioners as a
middle ground \citep{matveeva2025dynamic}. Yet the structure these
approximations discard is not negligible: transformer Hessians exhibit strong
heterogeneity across parameter blocks \citep{zhang2024transformers}, and the
prevailing near-block-diagonal picture is only approximate, with theory
attributing it to architectural and output-dimension effects rather than to
genuinely vanishing cross-layer terms \citep{dong2025towards}. No existing
estimator captures this cross-layer curvature at billion-parameter scale.

\paragraph{Curvature-guided compression.}
Second-order sensitivity underlies much of modern model compression. The
optimal brain compression framework and its LLM-scale successor use the
Hessian of a layer-wise reconstruction loss to decide which weights to prune
or how to round them during quantization \citep{frantar2022optimal,
frantar2022gptq}, and related sensitivity estimates drive layerwise
mixed-precision bit allocation \citep{zhao2026coopq}. In low-rank
compression, Fisher information reweights the decomposition toward
task-sensitive directions, progressing from diagonal approximations
\citep{hsu2022language} to a Kronecker-factored approximation of the
observed FIM within each layer \citep{chekalina2025generalized}. Curvature
further informs recovery after compression via Hessian-guided zeroth-order
fine-tuning \citep{zhao2025second}. Crucially, all of these methods treat
each layer independently, even though compression errors demonstrably
propagate and interact across layers \citep{arai2026quantization} -
precisely the structure a layer-local curvature estimate cannot see.

\paragraph{Our positioning.}
Across optimization and compression, a consistent pattern emerges: scalable
methods estimate sensitivity independently per layer, and compression
pipelines allocate precision, sparsity, or rank from layer-local proxies,
even though errors demonstrably interact across layers
\citep{arai2026quantization}. The closest work to ours, GFWSVD
\citep{chekalina2025generalized}, captures intra-layer parameter correlations
through a Kronecker-factored Fisher but remains a per-layer estimator tied to
a single compression task. We extend Kronecker-based Fisher approximation to
the \emph{cross-layer} setting, yielding a tractable estimate of inter-layer
curvature for billion-parameter models, and show that a single such estimate
transfers across compression problems: quantization,
sparsification,
as well as predicting recovery under fine-tuning.

\section{Hessian Kronecker Approximation}
\label{sec:methodology}

Let $W_1, \dots, W_k$ be the weight matrices of a deep neural network. We collect their parameters into a single vector $w = [w_1, w_2, \dots, w_k] \in \mathbb{R}^{d}$, where $w_i = \mathrm{vec}(W_i)$, $\mathrm{vec}(\cdot)$ denotes the vectorization operation, and $d$ is the total number of parameters. Let $g(w; x) \in \mathbb{R}^{d}$ denote the gradient of the loss with respect to $w$ on a data sample $x$. We approximate the Hessian $H(w)$ of the loss by the empirical Fisher matrix $J(w)$, defined as the expected outer product of the gradients over the data distribution. 
This approximation holds under the assumption that the model weights lie near a local optimum of the loss, a condition that is typically satisfied by converged pre-trained models
\begin{equation}
H(w) \approx J(w) = \mathbb{E}_{x}\!\left[g(w; x)\, g(w; x)^{\top}\right] \in \mathbb{R}^{d \times d}.
\label{hess_and_fish}
\end{equation}

As can be observed, the number of parameters in both the Hessian and the Fisher information matrix scales quadratically with the number of model parameters. Consequently, storing and manipulating full matrices becomes infeasible even for moderately sized neural networks, due to prohibitive memory and computational requirements. To avoid full matrix construction we propose our Kronecker-based Fisher matrix approximation.

We first vectorize the Fisher matrix \eqref{hess_and_fish} and express it in terms of Kronecker products:
\begin{equation}
\begin{aligned}
    \mathrm{vec}(J(w)) &= \mathbb{E}_x[\mathrm{vec}(g\,g^{\top})] = \mathbb{E}_x[g \otimes g] \in \mathbb{R}^{d^2},
\end{aligned}
\label{eq:first}
\end{equation}
where $\otimes$ denotes the Kronecker product and we write $g = g(w; x)$ for brevity. Reshaping the gradient into a matrix $G \in \mathbb{R}^{n \times m}$ with $g = \mathrm{vec}(G)$, equation \eqref{eq:first} becomes $\mathbb{E}_x[g \otimes g] = \mathbb{E}_x[\mathrm{vec}(G) \otimes \mathrm{vec}(G)]$. The Kronecker product of vectorizations can be converted into the vectorization of a Kronecker product using the symmetric permutation matrix \mbox{$P = I_n \otimes K_{nm} \otimes I_m = P^{\top} \in \mathbb{R}^{n^2m^2 \times n^2m^2}$}, where $K_{nm}$ is the commutation matrix:
\begin{equation}
\begin{aligned}
&\mathbb{E}_x[\mathrm{vec}(G) \otimes \mathrm{vec}(G)] = \mathbb{E}_x[P^\top \mathrm{vec}(G \otimes G)] \\
&= P^\top \mathrm{vec}\!\left(\mathbb{E}_x[G \otimes G]\right) = P^\top \mathrm{vec}\Big(\sum_{i=1}^{R}\sigma_i u_i v_i^\top\Big),
\end{aligned}
\label{eq:second}
\end{equation}
where the last equality uses the full-rank SVD \citep{golub1971singular} of $\mathbb{E}_x[G \otimes G]$ with rank $R$, singular values $\sigma_i$, and singular vectors $u_i \in \mathbb{R}^{n^2}$, $v_i \in \mathbb{R}^{m^2}$. Since $u_i$ and $v_i$ are themselves vectorizations of matrices $U_i \in \mathbb{R}^{n \times n}$ and $V_i \in \mathbb{R}^{m \times m}$, we can rewrite \eqref{eq:second} as
\begin{equation}
\begin{aligned}
     P^\top \mathrm{vec}\Big(\sum_{i=1}^{R}\sigma_i u_i v_i^\top\Big) &= P^\top \sum_{i=1}^{R} \sigma_i\, \mathrm{vec}(U_i) \otimes \mathrm{vec}(V_i) \\
     &= \sum_{i=1}^{R} \sigma_i\, \mathrm{vec}(U_i \otimes V_i) = \mathrm{vec}(J(w)),
\end{aligned}
\label{eq:third}
\end{equation}
where the second equality uses the identity \mbox{$\mathrm{vec}(U_i \otimes V_i) = (I_n \otimes K_{nm} \otimes I_m)(\mathrm{vec}(U_i) \otimes \mathrm{vec}(V_i))$}. 
Equation \eqref{eq:third} yields an exact Kronecker decomposition of the Fisher matrix, $J(w) = \sum_{i=1}^{R} \sigma_i\, U_i \otimes V_i$, which we approximate by truncating the sum to the $r < R$ largest singular values:
\begin{equation}
J(w) \approx \overline{J}(w) = \sum_{i=1}^{r} \sigma_i\, U_i \otimes V_i.
\label{lowrank}
\end{equation}
By the Eckart--Young theorem, this truncation is the best rank-$r$ approximation of $\mathbb{E}_x[G \otimes G]$ in the Frobenius norm.


Computing the decomposition \eqref{eq:third} directly is infeasible, since $\mathbb{E}_x[G \otimes G] \in \mathbb{R}^{n^2 \times m^2}$ is large. We therefore never form this matrix explicitly and instead obtain its leading singular values and vectors with an implicitly restarted Arnoldi method \citep{lehoucq1998arpack}, which accesses the matrix only through products with vectors. Exploiting the identity $(G \otimes G)\,\mathrm{vec}(V) = \mathrm{vec}(G V G^{\top})$, these products reduce to small dense multiplications with the reshaped gradients:
\begin{equation}
\begin{aligned}
    & \mathrm{matvec}(v) = \frac{1}{B} \sum_{b=1}^{B} \mathrm{vec}(G_b V G_b^{\top}), \\
    & \mathrm{rmatvec}(u) = \frac{1}{B} \sum_{b=1}^{B} \mathrm{vec}(G_b^{\top} U G_b),
\end{aligned}
\label{eq:compute}
\end{equation}
where $G_b \in \mathbb{R}^{n \times m}$ is the reshaped gradient accumulated over batch $b$, and $u = \mathrm{vec}(U)$ with $U \in \mathbb{R}^{n \times n}$ and $v = \mathrm{vec}(V)$ with $V \in \mathbb{R}^{m \times m}$ correspond to the two Kronecker factors. 



To improve the approximation further, we compute the diagonal of the Fisher matrix exactly,
\mbox{$\mathrm{diag}(J(w)) = \mathbb{E}_x[g \odot g]$}, where $\odot$ denotes the element-wise
product, and substitute it into the low-rank decomposition. The final approximation is therefore
$\overline{J}(w) = \sum_{i=1}^{r} \sigma_i\, U_i \otimes V_i$ with its diagonal replaced by
$\mathbb{E}_x[g \odot g]$, which is exact.
Table~\ref{tab:r2_vs_rank} shows that significantly improves the approximation.

\paragraph{Compressed Visualization}
Storing every entry of $\overline{J}(w)$ for visualization would require $O(d^2)$ memory, as much
as the full Fisher matrix. We therefore display a \textit{compressed} form, in which each
$m \times m$ block of the matrix is summarized by a single value, yielding an $n \times n$ image:
\begin{equation}
\begin{aligned}
&J_{\mathrm{vis}}(w) = \sum_{i=1}^{r} \sigma_i\, U_i \cdot \Big( \frac{1}{m^2} \sum_{p,q=1}^{m} (V_i)_{pq}\Big), \\
&\big(J_{\mathrm{vis}}(w)\big)_{jj} = \frac{1}{m} \sum_{t=(j-1)m+1}^{jm} \mathbb{E}_x\big[g_p^2\big].
\end{aligned}
\label{eq:visual}
\end{equation}
The off-diagonal entries of $J_{\mathrm{vis}}(w)$ are the mean values of the corresponding blocks,
which follows from the block structure $(U_i \otimes V_i)_{jk} = (U_i)_{jk} V_i$, while the diagonal
entries average the exact diagonal of the Fisher matrix within each block. The resulting image
requires only $O(n^2)$ memory, and the Kronecker factors it is built from require
$O(r(n^2 + m^2))$, which for $n \approx m \approx \sqrt{d}$ is comparable to the size of the model
itself for small $r$, as opposed to $O(d^2)$ for the full matrix.


\paragraph{Computational Complexity}
We now estimate the time and memory requirements of our method for a model, or a part of it, with
$d = nm$ parameters, using $B$ batches of $T$ tokens each. A backward pass costs $O(Td)$ per batch,
so computing the gradients takes $O(BTd)$; accumulating the exact diagonal from the per-batch
gradients adds $O(Bd)$. A single iteration of the Arnoldi method requires one product of the form
\eqref{eq:compute} per batch, that is $O(B(n^2m + nm^2)) = O(Bd(n+m))$ operations, and assembling the
compressed visualization from the resulting factors takes $O(r(n^2+m^2))$. With $N_A$ Arnoldi
iterations, the total time complexity is
\begin{multline*}
O\big(BTd + Bd + N_A B d (n+m) + r(n^2+m^2)\big) \\
= O\big(Bd\,(T + N_A (n+m))\big),
\end{multline*}
which for a balanced factorization $n \approx m \approx \sqrt{d}$ becomes
$O(Bd(T + N_A\sqrt{d}))$. The memory footprint consists of the gradient statistics, the diagonal,
and the $r$ pairs of Kronecker factors, $O(d + r(n^2+m^2)) = O(rd)$, which is linear in the number
of parameters for a fixed rank $r$, in contrast to the $O(d^2)$ required to store the Fisher matrix
explicitly.
We provide the end-to-end Hessian approximation construction times in Section~\ref{sec:hessian_computation_time}.

\section{Experiments}
\label{sec:experiments}
\subsection{Validating the Hessian Approximation}
\label{sec:hessian_validation}

Following \citet{dong2025towards}, we first validate our Hessian approximation on a small model where the true Hessian can be computed exactly: a two-layer perceptron trained with binary cross-entropy loss and the AdamW optimizer \citep{loshchilov2017decoupled} on a synthetic dataset of $500$ Gaussian clusters. The input and output dimensions are $500$ and the hidden dimension is $8$, for a total of $8$k parameters. 

\begin{figure}[th]
    \centering
    \includegraphics[width=1\linewidth]{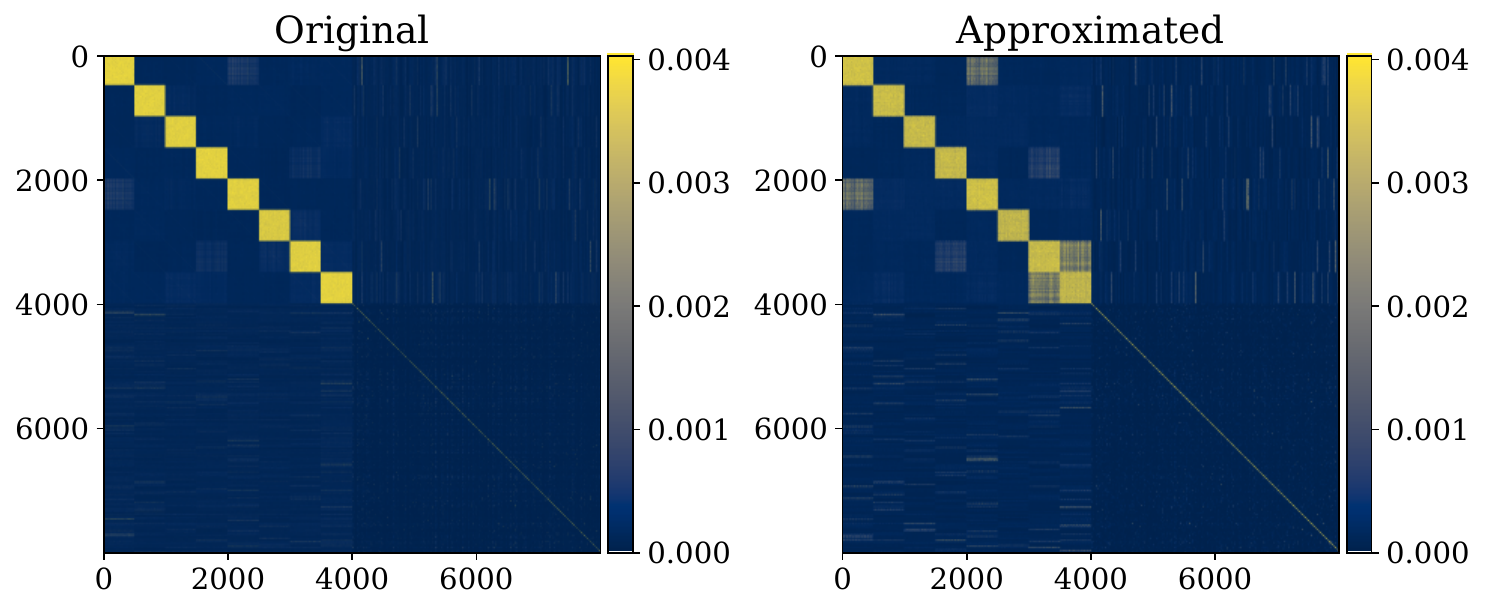}
    \caption{The true Hessian $H$ of the two-layer perceptron and its rank-$16$ Kronecker-based approximation $\overline{J}$. Axes denote the parameter index.}
    \label{fig:small_hess}
\end{figure}

As Figure~\ref{fig:small_hess} shows, the Kronecker-based approximation $\overline{J}$ recovers the structure of the true Hessian $H$. To quantify the agreement, we compute the coefficient of determination
$R^2(H, \overline{J}) = 1 - \frac{\|H - \overline{J}\|_F^2}{\|H\|_F^2}$,
which reaches $42.3\%$ with the explicit diagonal computation from \eqref{eq:visual} and $29.9\%$ without it. Table~\ref{tab:r2_vs_rank} details how $R^2$ varies with the rank of our Fisher matrix approximation: the explicit diagonal consistently improves the fit, with the largest relative gains at low ranks.

\begin{table}[h]
\centering
\begin{tabular}{l c c c c c}
\toprule
rank $r$ & 1 & 2 & 4 & 8 & 16 \\
\midrule
w/o diag & 3.2 & 9.8 & 15.6 & 24.5 & 29.9 \\
w/ diag  & 17.8 & 26.3 & 32.7 & 36.1 & 42.3 \\
\bottomrule
\end{tabular}
\caption{$R^2$ score (\%) between the true Hessian and our approximation for different ranks $r$, with and without explicit diagonal computation.}
\label{tab:r2_vs_rank}
\end{table}

\begin{figure*}[t]
\refstepcounter{figure}
\centering

\begin{minipage}[t]{0.32\textwidth}
\vspace{0pt}
\centering
\refstepcounter{subfigure}\label{fig:hessian_opt}
\includegraphics[width=\linewidth]{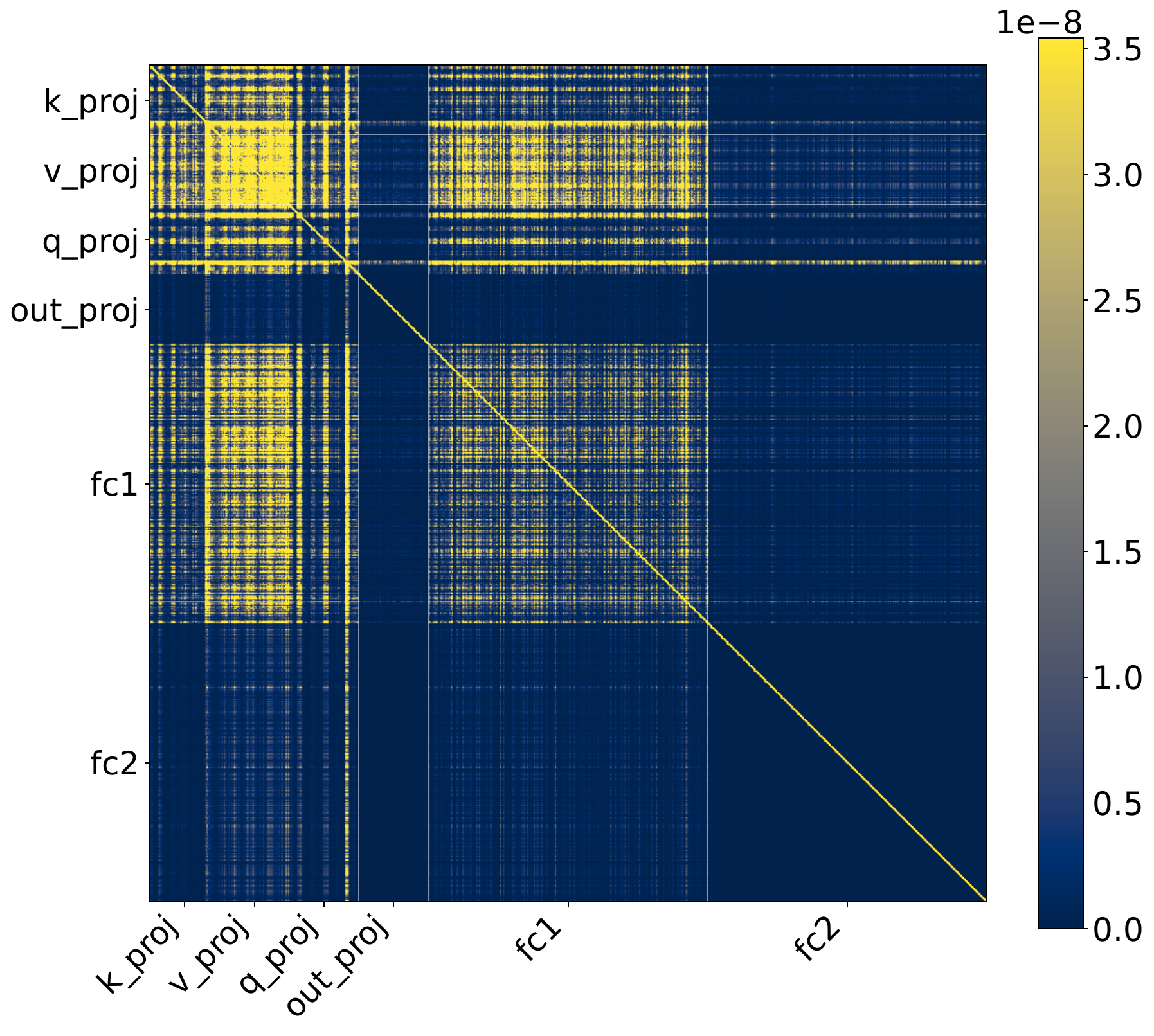}
\caption*{(\thesubfigure)\ OPT-350M}
\end{minipage}\hfill
\begin{minipage}[t]{0.32\textwidth}
\vspace{0pt}
\centering
\refstepcounter{subfigure}\label{fig:hessian_qwen}
\includegraphics[width=\linewidth]{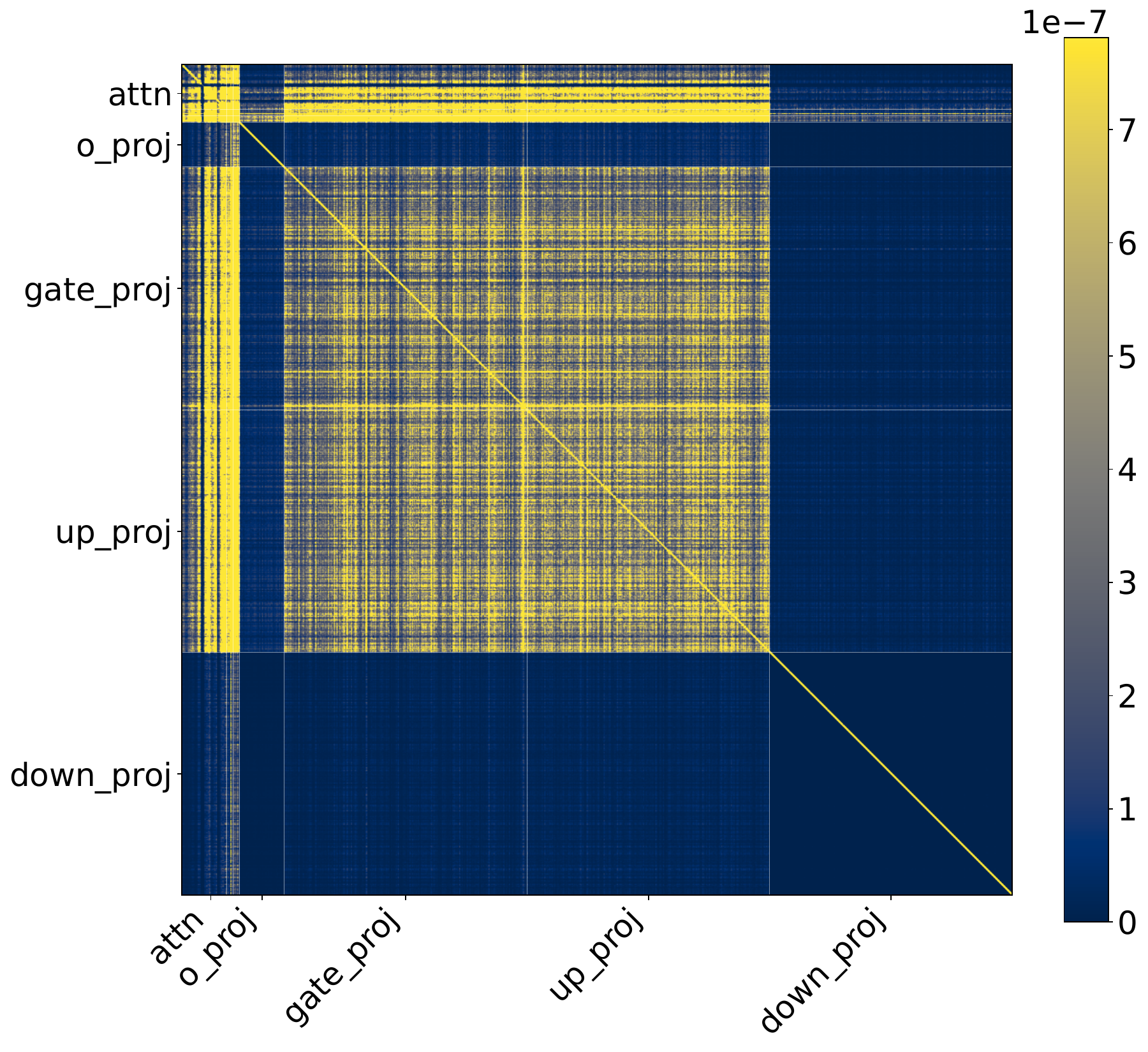}
\caption*{(\thesubfigure)\ Qwen2.5-0.5B}
\end{minipage}\hfill
\begin{minipage}[t]{0.32\textwidth}
\vspace{0pt}
\centering
\refstepcounter{subfigure}\label{fig:hessian_olmo}
\includegraphics[width=\linewidth]{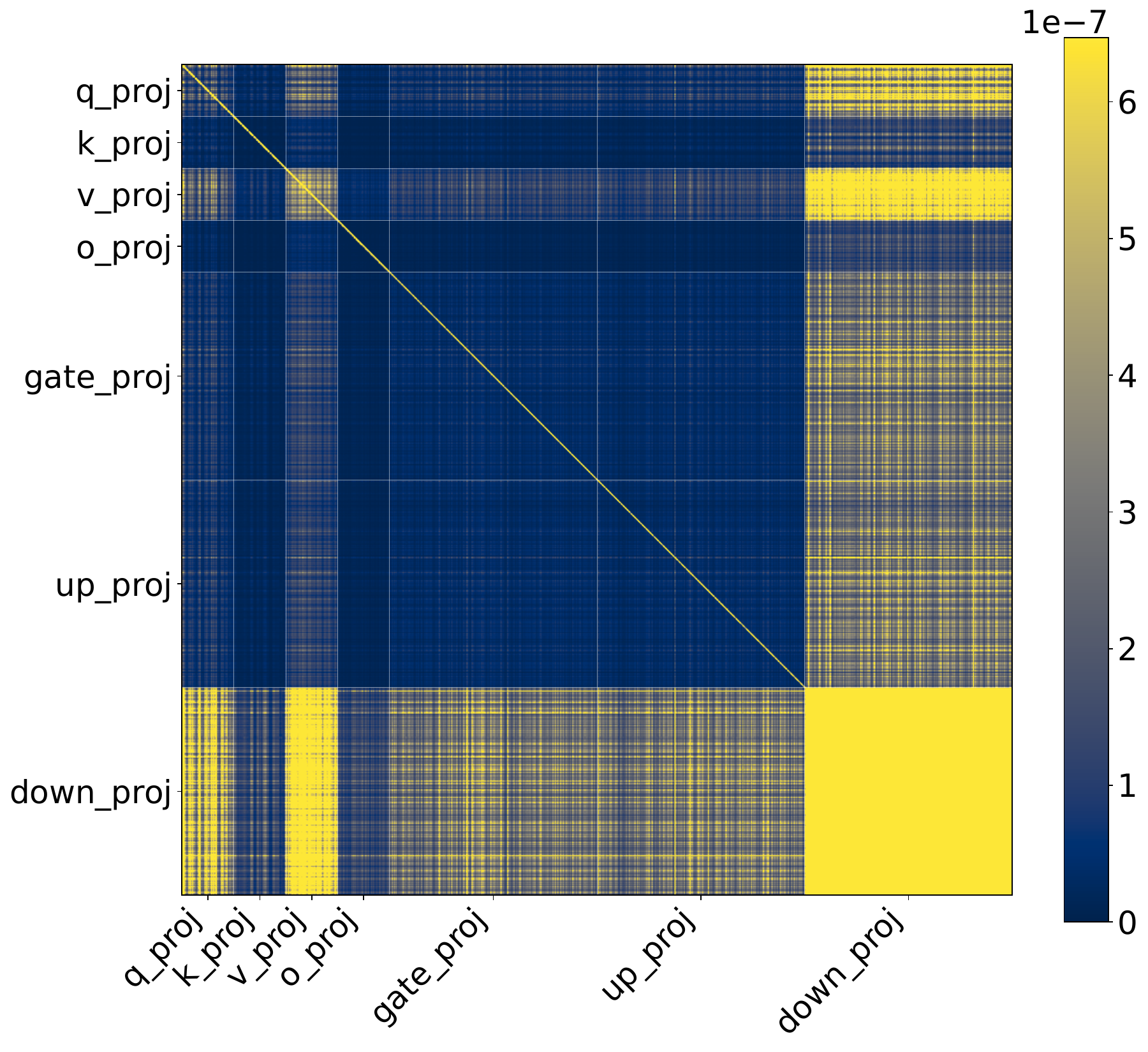}
\caption*{(\thesubfigure)\ OLMo2-1B}
\end{minipage}

\addtocounter{figure}{-1}
\caption{Kronecker-Fisher approximation of the Hessian (Fisher Information) for one Transformer block of OPT-350M, Qwen2.5-0.5B, and OLMo2-1B. For the Qwen2.5 architecture, where the $K$- and $V$-projections are very small and most of the Transformer block parameters belong to the MLP linear layers, we group the entire attention mechanism ($Q$, $K$, $V$) under a single label \texttt{attn} rather than labeling each projection individually. We also observed that the Hessian structure of Qwen2.5-7B is the same as Qwen2.5-0.5B.}
\label{fig:hessian_comparison}
\end{figure*}

\bigskip
All subsequent experiments are conducted on four large language models spanning different families and sizes: OPT-350M \citep{zhang2022opt}, Qwen2-0.5B \citep{qwen2}, OLMo2-1B \citep{olmo20242olmo2furious}, and Qwen2.5-7B \citep{qwen2}. For evaluation, we use text from the WikiText2 corpus \citep{merity2016pointer}. We first construct approximate Hessians over several batches of WikiText2, computing gradients over groups of full Transformer blocks with $N$ parameters and visualizing the approximation in the compressed form \eqref{eq:visual}, where we set $n \approx \sqrt{N}$, $n \in \mathbb{N}$. 

In the main paper, we restrict the Hessian visualizations to the level of individual Transformer blocks, as the full matrix is too large to display legibly even in the compressed form \eqref{eq:visual}. 
Full-model and multi-block visualizations are provided in the Supplementary.

\begin{figure*}
    \centering
    \includegraphics[width=1\linewidth]{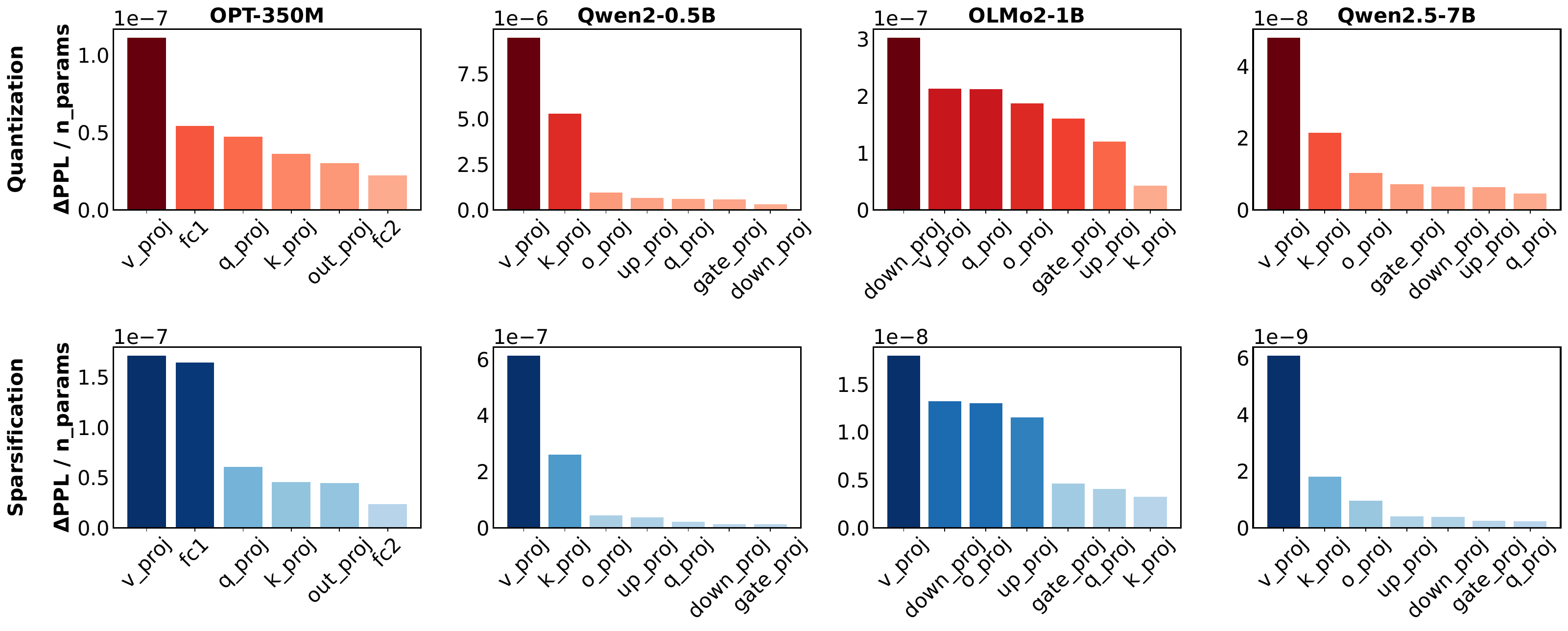}
    \caption{Increase in Perplexity scores following 4-bit layer quantization (top row) and 50\%-sparsification (bottom row) of various Transformer block components in the middle layers (excluding first 3 and last 3 transformer blocks) of the OPT-350M, Qwen2-0.5B, OLMo2-1B and Qwen 2.5-7B models. The scores are normalized by the number of parameters in the quantized layers to compare the relative effects of quantization across layers with different parameter counts. The layers are sorted in descending order based on the value of perplexity change.}
    \label{fig:quantandsparse}
\end{figure*}

As our next step, we test whether the approximate Hessian predicts layer sensitivity to compression: first within groups of layers of the same type (query, value, up-projection, etc.; Figure~\ref{fig:quantandsparse}), and second across layer types, by compressing layers of different types (Figure~\ref{fig:pair_quant}). 

\subsection{Compression within a Single Layer Type}
\label{sec:single_layer_type}

To connect our Kronecker-based Fisher matrix approximation to practical model compression, we study how sensitive individual layer types are to weight corruption. In this section, we compress only one layer type at a time (e.g., only the $V$-projections or only the upscale layers); interactions between different layer types within the same Transformer block are studied in Section~\ref{sec:mult_layer_type}. 
In both settings, we compress the corresponding layers in all Transformer blocks except the first and last three, and measure the resulting increase in perplexity on WikiText2. We exclude these boundary blocks because they often exhibit out-of-distribution behavior, and our Hessian approximations indicate weaker correspondence between their parameters (see Supplementary).

\paragraph{Quantization.}

We first quantize the selected layers with $4$-bit uniform quantization. As shown in the first row in Figure~\ref{fig:quantandsparse}, the $V$-projection causes the highest perplexity increase in all models except OLMo2-1B, where the downscale projection is the most vulnerable. These results are consistent with the corresponding Hessian approximations (Figure~\ref{fig:hessian_comparison}): the largest values appear on the $V$-projection for the OPT (Figure~\ref{fig:hessian_opt}) and Qwen (Figure~\ref{fig:hessian_qwen}) models, and on the downscale projection for OLMo2 (Figure~\ref{fig:hessian_olmo}).

The remaining sensitive layers are model-specific, but in each case they match the layers our approximation highlights: quantizing the upscale layer causes a significant quality drop in OPT-350M, consistent with its high values in Figure~\ref{fig:hessian_opt}; the $K$-projection is vulnerable in Qwen2-0.5B and Qwen2.5-7B, in line with Figure~\ref{fig:hessian_qwen}; and in OLMo2-1B, quantizing the $V$- and $Q$-projections and the downscale layer leads to a substantial perplexity increase, matching the largest values of our Kronecker-Fisher approximation in Figure~\ref{fig:hessian_olmo}.

\paragraph{Sparsification.}

We observe the same pattern under sparsification with ratio $0.5$ (second row in Figure~\ref{fig:quantandsparse}). The $V$-projections exhibit the highest sensitivity in all models, in agreement with the Fisher matrix results in Figure~\ref{fig:hessian_comparison}. 
The only model whose vulnerability ranking differs between quantization and sparsification is OLMo2: the $V$-projection is the most sensitive to sparsification, with the downscale projection in second place, whereas for quantization their order is reversed. Both layers exhibit the highest values in the Hessian approximation for OLMo2 (Figure~\ref{fig:hessian_olmo}), with the downscale projection being the more pronounced of the two. This discrepancy may stem from our normalization scheme: since we normalize the perplexity increase by the number of parameters in the compressed layer, and the $V$-projection is much smaller than the downscale projection, the per-parameter effect of the $V$-projection is amplified.

Overall, across both compression schemes, the layers assigned high values by our Fisher matrix approximation are exactly those whose corruption degrades model quality the most, with the $V$-projection being the most vulnerable layer type throughout.

\subsection{Inter-Layer Quantization and Sparsification}
\label{sec:mult_layer_type}

\begin{figure}
    \centering
    \includegraphics[width=1\linewidth]{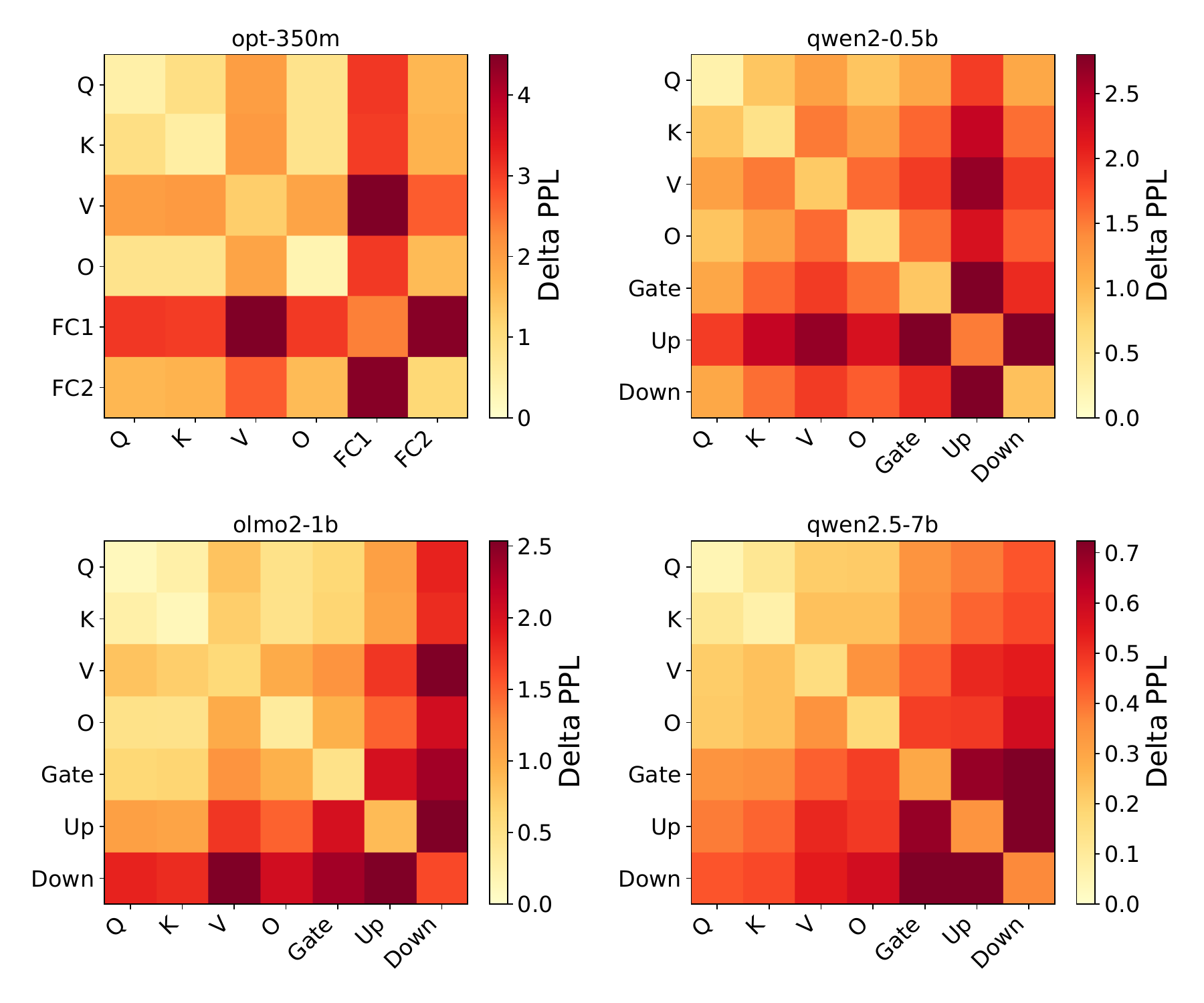}
    \caption{The Perplexity growth following quantization in 4 bits the different pairs of layers in Transformer blocks in the middle of the LLM models. The value on crossing the row $i$ and the column $j$ is the result of PPL growth after quantization of layers $i$ and $j$.}
    \label{fig:pair_quant}
\end{figure}

\begin{figure}
    \centering
    \includegraphics[width=1\linewidth]{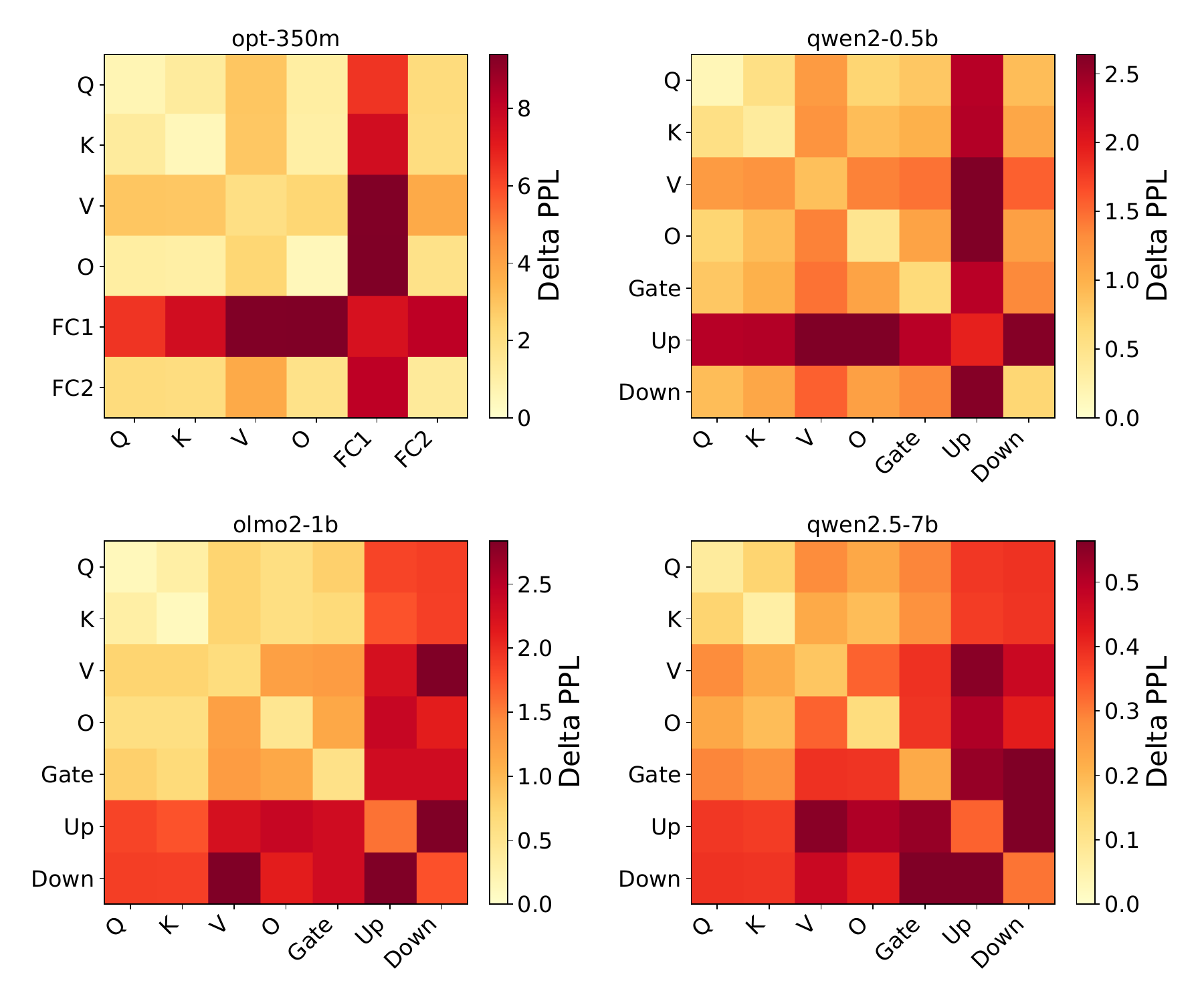}
    \caption{The Perplexity growth following the sparsification with ratio $0.5$ the different pairs of layers in Transformer blocks in the middle of the LLM models. The value on crossing the row $i$ and the column $j$ is the result of PPL growth after sparsification of layers $i$ and $j$.}
    \label{fig:pair_sparse}
\end{figure}

To study inter-layer interactions in the Fisher matrices, we also quantize and sparsify pairs of layers, such as the $V$-projection together with the upscale layer. As in Section~\ref{sec:single_layer_type}, we apply $4$-bit quantization and sparsification with ratio $0.5$ to all Transformer blocks except the first and last three, and evaluate the corrupted models on WikiText2 (Figures~\ref{fig:pair_quant} and \ref{fig:pair_sparse}). 

The greatest perplexity increase occurs for pairs involving the $V$-projection and for pairs combining the upscale and downscale layers. Note that Figures~\ref{fig:pair_quant} and \ref{fig:pair_sparse} are not normalized by the number of parameters, so pairs of large layers naturally appear brighter. What matters is which pairs stand out beyond this size effect, and these are precisely the pairs highlighted by our Hessian approximations: the $V$--FC1 pair for OPT, consistent with Figure~\ref{fig:hessian_opt}; the $V$--upscale pair for Qwen2.5, consistent with Figure~\ref{fig:hessian_qwen}; and the $V$--downscale pair for OLMo2, consistent with Figure~\ref{fig:hessian_olmo}. Therefore, the off-diagonal structure of our Kronecker-based Fisher matrix approximation predicts not only which individual layers are sensitive to compression, but also which pairs of layers interact most strongly when corrupted together.

A natural concern is that the highlighted pairs in Figures~\ref{fig:pair_quant} and \ref{fig:pair_sparse}, such as $V$--upscale, merely reflect the accumulation of two independent errors: if quantizing $V$ and quantizing FC1 each degrade the model on their own, their combination would stand out even without any genuine interaction between the layers. To rule out this explanation, we compute the interaction difference $D_I$ for pairs of layers. Formally, for layers $P$ and $Q$, \mbox{$D_I(P, Q) = \Delta(\{P, Q\}) - \Delta(\{P\}) - \Delta(\{Q\})$}, where $\Delta(S)$ denotes the perplexity increase after corrupting the layers in set $S$. By construction, $D_I$ removes the additive contribution of each layer and isolates the excess degradation caused by corrupting the two layers jointly. The largest interaction differences between attention and MLP layers occur precisely for the pairs with the highest values in our Hessian approximations (Figure~\ref{fig:hessian_comparison}), which are the pairs involving the $V$-projection. This confirms that the prominence of these pairs is a genuine interaction effect, captured by the off-diagonal structure of the Fisher matrix, rather than an artifact of cumulative independent errors.

An exception to the otherwise strong agreement between our Fisher matrix approximation and the empirical results is the $O$-projection. While the corresponding regions of the approximated Hessians show comparatively low values in all models, pairs involving the $O$-projection, particularly $O$--upscale, cause a notable perplexity increase under both quantization and sparsification (Figures~\ref{fig:pair_quant} and \ref{fig:pair_sparse}). We hypothesize that the discrepancy arises from the local nature of the Fisher approximation. The $O$-projection writes the attention output directly into the residual stream, which is known to carry a small number of high-magnitude outlier channels that the MLP layers rely on. Near the trained optimum, gradients through the $O$-projection are small, yielding low Fisher values; however, the finite perturbations introduced by $4$-bit quantization or $50\%$ sparsification disturb these outlier channels, causing damage that a local quadratic model cannot capture. 
We discuss the further in the Future Research (Section~\ref{sec:conclusion}).

We
additionally evaluate the corrupted models on five zero-shot benchmarks (PIQA, WinoGrande,
HellaSwag, ARC-Easy, ARC-Challenge), where the resulting layer rankings closely match those obtained
from perplexity. 
Full per-task results are reported in Supplementary.

\subsection{Finetuning}

In modern model optimization pipelines, compressed models are typically fine-tuned to recover acceptable performance \citep{liao-etal-2024-apiq}. We investigate whether our Fisher matrix approximation can predict which layers are most effective to fine-tune. To this end, we first corrupt all MLP layers (upscale, gate, and downscale) of each language model with $4$-bit quantization or sparsification with ratio $0.5$, and then fine-tune individual attention layers, one at a time (Figure~\ref{finetune}). Fine-tuning the $V$-projection recovers the most performance, whereas fine-tuning the $K$- or $Q$-projections yields little improvement. This mirrors the off-diagonal structure of our Hessian approximations, where the $V$-projection exhibits the strongest coupling with the FFN layers.

\begin{figure}
    \centering
    \includegraphics[width=1\linewidth]{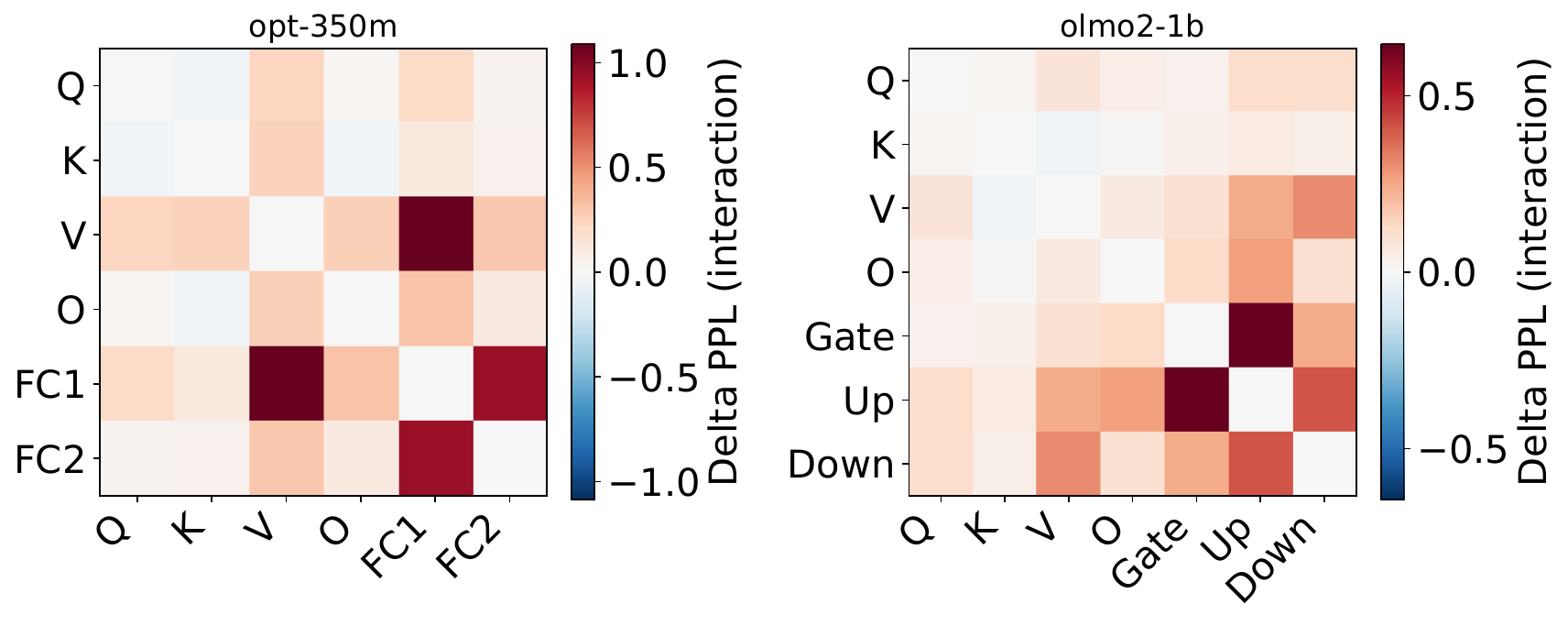}
    \caption{Interaction difference $D_I$ for $4$-bit quantization in OPT-350M and OLMo2-1B. The cell at row $i$ and column $j$ shows $D_I(i, j)$: the perplexity increase after jointly quantizing layer types $i$ and $j$, minus the sum of the perplexity increases after quantizing each layer type individually. Positive values indicate that the joint corruption is more damaging than the two individual corruptions combined.}
    \label{inter_quant}
\end{figure}

\begin{figure}
    \centering
    \includegraphics[width=1.0\linewidth]{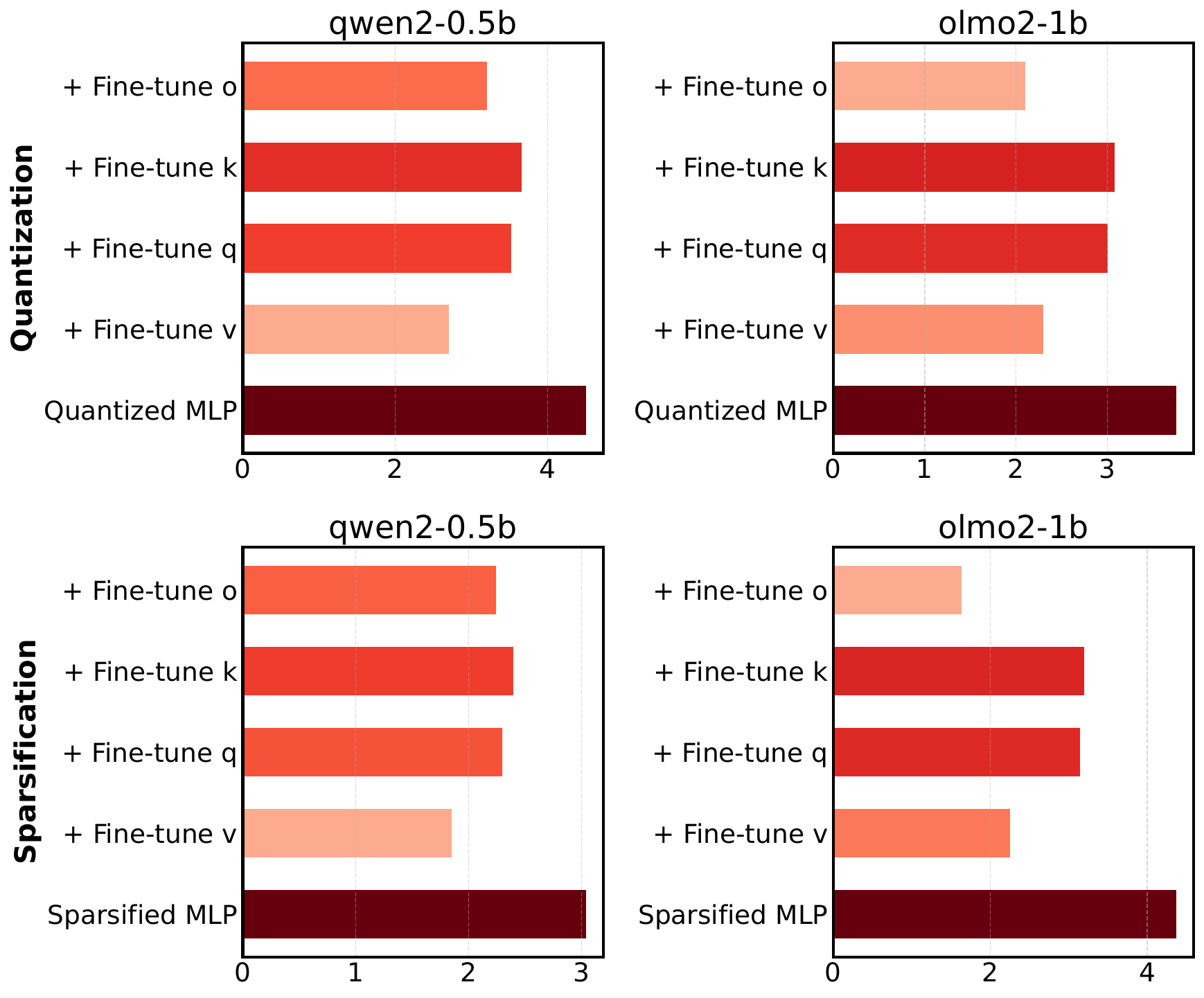}
    \caption{The Perplexity increase after corrupting all FFN layers, before and after fine-tuning individual attention layers. The top row shows results for $4$-bit quantization and the bottom row for sparsification with ratio $0.5$.}
    \label{finetune}
\end{figure}

\begin{figure}
    \centering
    \includegraphics[width=1.0\linewidth]{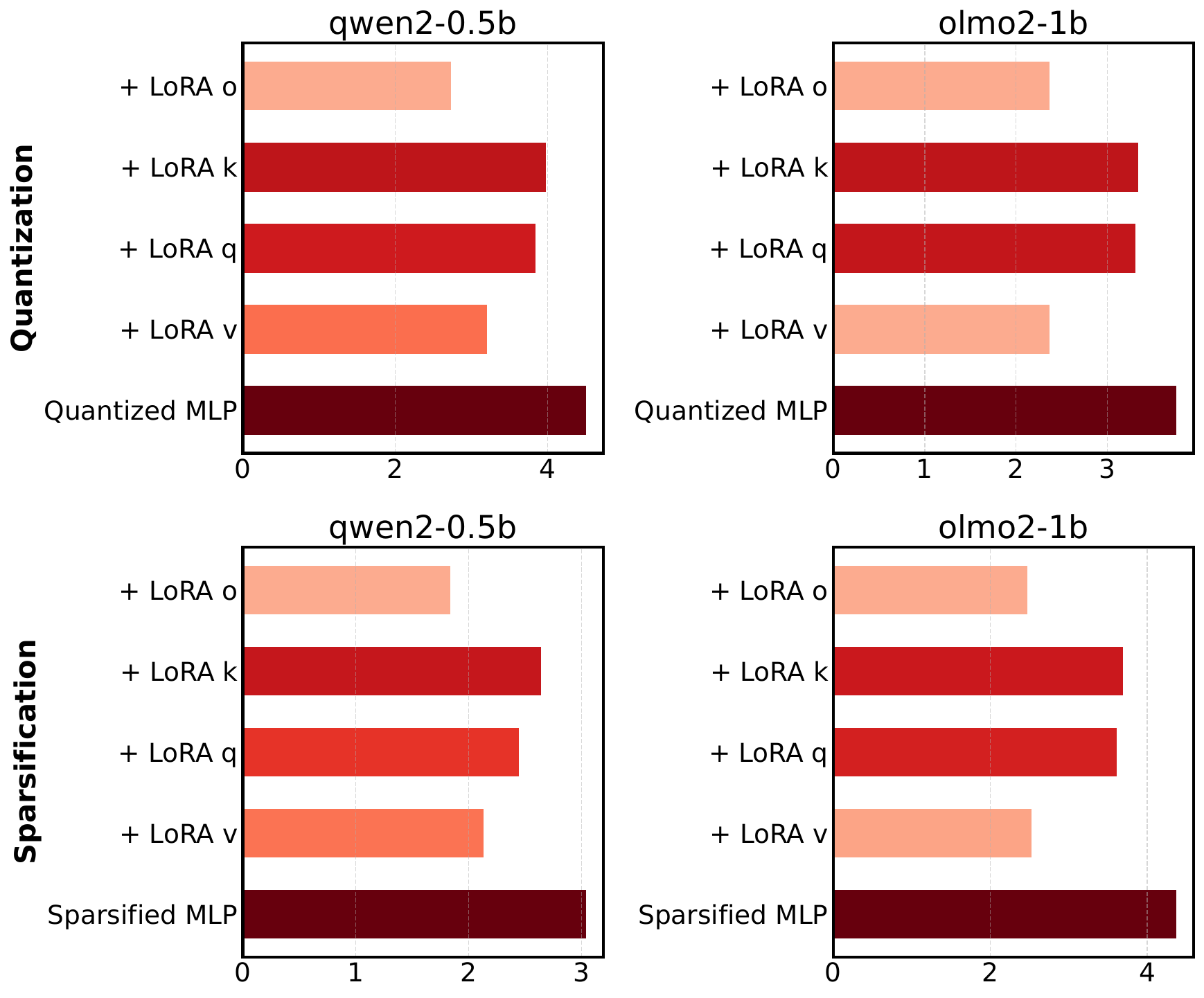}
    \caption{The Perplexity increase after FFN corruption and the its values after LoRA application to various Attention layers. The top rows demonstrates the experiments with 4-bit quantization and the bottom one with sparsification.}
    \label{lora}
\end{figure}

We additionally repeat this experiment with low-rank adapters (LoRA) \citep{hu2022lora} in place of full fine-tuning. LoRA adds a trainable low-rank correction to a frozen weight matrix: $W_{\mathrm{new}} = W + AB^T$, where $W \in \mathbb{R}^{n \times m}$ is frozen and $A \in \mathbb{R}^{n \times r}$, $B \in \mathbb{R}^{m \times r}$ are trainable matrices of rank $r$. As shown in Figure~\ref{lora}, applying the adapter to the $V$-projection is again substantially more effective than applying it to the $Q$- or $K$-projections, in agreement with our Kronecker-based Hessian approximation of the Transformer block (Figures~\ref{fig:hessian_opt},~\ref{fig:hessian_olmo},~\ref{fig:hessian_qwen}). 

Overall, the fine-tuning experiments show that adapting layers with high off-diagonal Hessian values with respect to the corrupted layers recovers substantially more performance than adapting weakly coupled layers. This further confirms that the off-diagonal values of our approximation reflect genuine inter-layer dependencies in large language models. Our method (Section~\ref{sec:methodology}) therefore offers a practical criterion for selecting which layers to fine-tune in order to compensate for compression errors.

\subsection{Hessian Computation Time}
\label{sec:hessian_computation_time}
To support our claim that the proposed approximation is tractable at scale, we report the wall-clock
time required to construct it. We use a rank-1 approximation, already sufficient to reveal the
Hessian structure, with $20$ batches of $10$ WikiText2 sequences on a single NVIDIA H100 GPU. As
Table~\ref{tab:full-model-hessian-construction-time} shows, constructing the approximation for an
entire model takes from a few minutes for OPT-125M to roughly $40$ minutes for OLMo2-1B, so the full
curvature structure of a billion-parameter model is obtainable within a single GPU-hour. Gradient
computation accounts for a decreasing share of the total as models grow, consistent with the
complexity analysis of Section~\ref{sec:methodology}, where the Arnoldi iterations dominate the
linear cost of backpropagation.

\begin{table}[t]
    \centering
    \caption{Full-model Hessian approximation construction time over all Transformer blocks, using
    a rank-1 approximation. Gradient computation time is included in the
    total.}
    \label{tab:full-model-hessian-construction-time}
    \begin{tabular}{lrr}
        \hline
        Model & Total time (s) & Gradient time (s) \\
        \hline
        OPT-125M   & 157.64  & 50.73  \\
        OPT-350M   & 1563.07 & 272.94  \\
        Qwen2-0.5B & 1090.61 & 200.09  \\
        OLMo2-1B   & 2428.25 & 218.64 \\
        \hline
    \end{tabular}
\end{table}

\section{Conclusion}
\label{sec:conclusion}
We introduced a scalable Kronecker-based Fisher approximation that enables practical Hessian analysis for billion-parameter language models, reducing memory complexity from quadratic to linear in model size while preserving cross-layer interaction information. Our method bridges the gap between theoretical curvature studies and real-world model optimization, providing the first direct empirical evidence of non-diagonal Hessian structure in large LLMs.

Across four model families, we consistently found that value projection layers exhibit the highest sensitivity and strongest cross-layer correlations, while other components show architecture-specific behavior. Extensive experiments on quantization, sparsification, inter-layer corruption, and fine-tuning demonstrate that our approximation strongly correlates with performance degradation and recovery, outperforming diagonal estimates and uncovering non-additive joint effects. These insights enable principled identification of fragile layers, guiding mixed‑precision compression, targeted fine‑tuning, and LoRA adaptation without the prohibitive cost of full Hessian computation. Our framework offers a practical, theoretically grounded tool for efficient model optimization, advancing both the understanding and compression of large-scale neural networks.

\paragraph{Future Work}

A promising direction for future work is to test whether the observed inter-layer interactions are mediated by outlier channels in the residual stream. Outlier-mitigation techniques such as AWQ \citep{lin2024awq} and rotation-based methods such as QuaRot \citep{ashkboos2024quarot} redistribute or remove these outliers. If the strong interactions we observe, such as the 
V--upscale pairs, arise because corruption in one layer damages the few high-magnitude channels the other depends on, then applying these techniques before compression should disproportionately reduce $D_I$ for exactly these pairs. The same experiment would clarify the 
O-projection discrepancy: a sharp drop in its vulnerability after outlier removal would indicate that this vulnerability stems from finite perturbations of outlier channels rather than from local curvature.

\bibliography{aaai2027}

@incollection{golub1971singular,
  title={Singular value decomposition and least squares solutions},
  author={Golub, Gene H and Reinsch, Christian},
  booktitle={Linear algebra},
  pages={134--151},
  year={1971},
  publisher={Springer}
}

@article{van2000ubiquitous,
  title     = {The ubiquitous {Kronecker} product},
  author    = {Van Loan, Charles F.},
  journal   = {Journal of Computational and Applied Mathematics},
  volume    = {123},
  number    = {1-2},
  pages     = {85--100},
  year      = {2000},
  publisher = {Elsevier}
}

@book{lehoucq1998arpack,
  title     = {ARPACK Users' Guide: Solution of Large-Scale Eigenvalue Problems with Implicitly Restarted {Arnoldi} Methods},
  author    = {Lehoucq, Richard B. and Sorensen, Danny C. and Yang, Chao},
  year      = {1998},
  publisher = {SIAM}
}

@book{izmailov2014newton,
  title     = {Newton-Type Methods for Optimization and Variational Problems},
  author    = {Izmailov, Alexey F. and Solodov, Mikhail V.},
  year      = {2014},
  publisher = {Springer}
}

@inproceedings{martens2010deep,
  title     = {Deep learning via {Hessian}-free optimization},
  author    = {Martens, James},
  booktitle = {Proceedings of the 27th International Conference on Machine Learning (ICML)},
  pages     = {735--742},
  year      = {2010}
}

@inproceedings{martens2015optimizing,
  title     = {Optimizing neural networks with {Kronecker}-factored approximate curvature},
  author    = {Martens, James and Grosse, Roger},
  booktitle = {International Conference on Machine Learning (ICML)},
  pages     = {2408--2417},
  year      = {2015},
  organization = {PMLR}
}

@article{zhang2017block,
  title     = {Block-diagonal {Hessian}-free optimization for training neural networks},
  author    = {Zhang, Huishuai and Xiong, Caiming and Bradbury, James and Socher, Richard},
  journal   = {arXiv preprint arXiv:1712.07296},
  year      = {2017}
}

@inproceedings{dangel2020modular,
  title     = {Modular block-diagonal curvature approximations for feedforward architectures},
  author    = {Dangel, Felix and Harmeling, Stefan and Hennig, Philipp},
  booktitle = {International Conference on Artificial Intelligence and Statistics (AISTATS)},
  pages     = {799--808},
  year      = {2020},
  organization = {PMLR}
}

@inproceedings{botev2017practical,
  title     = {Practical {Gauss-Newton} optimisation for deep learning},
  author    = {Botev, Aleksandar and Ritter, Hippolyt and Barber, David},
  booktitle = {International Conference on Machine Learning (ICML)},
  pages     = {557--565},
  year      = {2017},
  organization = {PMLR}
}

@article{naumov2017feedforward,
  title     = {Feedforward and recurrent neural networks backward propagation and {Hessian} in matrix form},
  author    = {Naumov, Maxim},
  journal   = {arXiv preprint arXiv:1709.06080},
  year      = {2017}
}

@techreport{collobert2004large,
  title     = {Large scale machine learning},
  author    = {Collobert, Ronan},
  institution = {IDIAP},
  number    = {RR-04-42},
  year      = {2004}
}

@inproceedings{kingma2014adam,
  title     = {Adam: A method for stochastic optimization},
  author    = {Kingma, Diederik P. and Ba, Jimmy},
  booktitle = {International Conference on Learning Representations (ICLR)},
  year      = {2015},
  note      = {arXiv:1412.6980}
}

@inproceedings{yao2021adahessian,
  title     = {{AdaHessian}: An adaptive second order optimizer for machine learning},
  author    = {Yao, Zhewei and Gholami, Amir and Shen, Sheng and Mustafa, Mustafa and Keutzer, Kurt and Mahoney, Michael},
  booktitle = {Proceedings of the AAAI Conference on Artificial Intelligence},
  volume    = {35},
  pages     = {10665--10673},
  year      = {2021}
}

@inproceedings{liu2024sophia,
  title     = {Sophia: A scalable stochastic second-order optimizer for language model pre-training},
  author    = {Liu, Hong and Li, Zhiyuan and Hall, David and Liang, Percy and Ma, Tengyu},
  booktitle = {International Conference on Learning Representations (ICLR)},
  year      = {2024},
  note      = {arXiv:2305.14342}
}

@article{das2024towards,
  title     = {Towards quantifying the preconditioning effect of {Adam}},
  author    = {Das, Rudrajit and Agarwal, Naman and Sanghavi, Sujay and Dhillon, Inderjit S.},
  journal   = {arXiv preprint arXiv:2402.07114},
  year      = {2024}
}

@inproceedings{zhang2025adam,
  title     = {Adam-mini: Use fewer learning rates to gain more},
  author    = {Zhang, Yushun and Chen, Congliang and Li, Ziniu and Ding, Tian and Wu, Chenwei and Kingma, Diederik P. and Ye, Yinyu and Luo, Zhi-Quan and Sun, Ruoyu},
  booktitle = {International Conference on Learning Representations (ICLR)},
  year      = {2025},
  note      = {arXiv:2406.16793}
}

@article{matveeva2025dynamic,
  title     = {Dynamic low-rank approximation of full-matrix preconditioner for training generalized linear models},
  author    = {Matveeva, Tatyana and Katrutsa, Aleksandr and Frolov, Evgeny},
  journal   = {arXiv preprint arXiv:2508.21106},
  year      = {2025}
}

@article{zhang2024transformers,
  title     = {Why transformers need {Adam}: A {Hessian} perspective},
  author    = {Zhang, Yushun and Chen, Congliang and Ding, Tian and Li, Ziniu and Sun, Ruoyu and Luo, Zhi-Quan},
  journal   = {Advances in Neural Information Processing Systems},
  volume    = {37},
  pages     = {131786--131823},
  year      = {2024}
}

@article{dong2025towards,
  title     = {Towards quantifying the {Hessian} structure of neural networks},
  author    = {Dong, Zhaorui and Zhang, Yushun and Yao, Jianfeng and Sun, Ruoyu},
  journal   = {arXiv preprint arXiv:2505.02809},
  year      = {2025}
}

@article{frantar2022optimal,
  title     = {Optimal brain compression: A framework for accurate post-training quantization and pruning},
  author    = {Frantar, Elias and Alistarh, Dan},
  journal   = {Advances in Neural Information Processing Systems},
  volume    = {35},
  pages     = {4475--4488},
  year      = {2022}
}

@inproceedings{frantar2022gptq,
  title     = {{GPTQ}: Accurate post-training quantization for generative pre-trained transformers},
  author    = {Frantar, Elias and Ashkboos, Saleh and Hoefler, Torsten and Alistarh, Dan},
  booktitle = {International Conference on Learning Representations (ICLR)},
  year      = {2023},
  note      = {arXiv:2210.17323}
}

@inproceedings{zhao2026coopq,
  title     = {{CoopQ}: Cooperative game inspired layerwise mixed precision quantization for {LLMs}},
  author    = {Zhao, Junchen and Derakhshan, Ali and Hyman, Jayden and Dong, Junhao and Jyothi, Sangeetha Abdu and Harris, Ian},
  booktitle = {Findings of the Association for Computational Linguistics: ACL 2026},
  pages     = {7566--7578},
  year      = {2026}
}

@article{arai2026quantization,
  title     = {Quantization error propagation: Revisiting layer-wise post-training quantization},
  author    = {Arai, Yamato and Ichikawa, Yuma},
  journal   = {Advances in Neural Information Processing Systems},
  volume    = {38},
  pages     = {151916--151951},
  year      = {2026}
}

@inproceedings{hsu2022language,
title={Language model compression with weighted low-rank factorization},
author={Yen-Chang Hsu and Ting Hua and Sungen Chang and Qian Lou and Yilin Shen and Hongxia Jin},
booktitle={International Conference on Learning Representations},
year={2022},
url={https://openreview.net/forum?id=uPv9Y3gmAI5}
}

@article{chekalina2025generalized,
  title     = {Generalized {Fisher}-weighted {SVD}: Scalable {Kronecker}-factored {Fisher} approximation for compressing large language models},
  author    = {Chekalina, Viktoriia and Moskovskiy, Daniil and Cherniuk, Daria and Kurkin, Maxim and Kuznetsov, Andrey and Frolov, Evgeny},
  journal   = {arXiv preprint arXiv:2505.17974},
  year      = {2025}
}

@inproceedings{zhao2025second,
  title     = {Second-order fine-tuning without pain for {LLMs}: A {Hessian} informed zeroth-order optimizer},
  author    = {Zhao, Yanjun and Dang, Sizhe and Ye, Haishan and Dai, Guang and Qian, Yi and Tsang, Ivor},
  booktitle = {International Conference on Learning Representations (ICLR)},
  year      = {2025}
}

@article{merity2016pointer,
  title     = {Pointer sentinel mixture models},
  author    = {Merity, Stephen and Xiong, Caiming and Bradbury, James and Socher, Richard},
  journal   = {arXiv preprint arXiv:1609.07843},
  year      = {2016}
}

@article{loshchilov2017decoupled,
  title     = {Decoupled weight decay regularization},
  author    = {Loshchilov, Ilya and Hutter, Frank},
  journal   = {arXiv preprint arXiv:1711.05101},
  year      = {2017}
}

@article{zhang2022opt,
  title     = {{OPT}: Open pre-trained transformer language models},
  author    = {Zhang, Susan and Roller, Stephen and Goyal, Naman and Artetxe, Mikel and Chen, Moya and Chen, Shuohui and Dewan, Christopher and Diab, Mona and Li, Xian and Lin, Xi Victoria and others},
  journal   = {arXiv preprint arXiv:2205.01068},
  year      = {2022}
}

@article{qwen2,
  title     = {Qwen2 technical report},
  author    = {Yang, An and Yang, Baosong and Hui, Binyuan and Zheng, Bo and Yu, Bowen and Zhou, Chang and Li, Chengpeng and Li, Chengyuan and Liu, Dayiheng and Huang, Fei and others},
  journal   = {arXiv preprint arXiv:2407.10671},
  year      = {2024}
}

@misc{olmo20242olmo2furious,
  title     = {2 {OLMo} 2 {Furious}},
  author    = {{Team OLMo} and Walsh, Pete and Soldaini, Luca and Groeneveld, Dirk and Lo, Kyle and Arora, Shane and Bhagia, Akshita and Gu, Yuling and Huang, Shengyi and Jordan, Matt and others},
  year      = {2024},
  eprint    = {2501.00656},
  archivePrefix = {arXiv},
  url       = {https://arxiv.org/abs/2501.00656}
}

@article{wu2025large,
  title     = {How large language models encode theory-of-mind: A study on sparse parameter patterns},
  author    = {Wu, Yuheng and Guo, Wentao and Liu, Zirui and Ji, Heng and Xu, Zhaozhuo and Zhang, Denghui},
  journal   = {npj Artificial Intelligence},
  volume    = {1},
  number    = {1},
  pages     = {20},
  year      = {2025}
}

@inproceedings{lin2024awq,
 author = {Lin, Ji and Tang, Jiaming and Tang, Haotian and Yang, Shang and Chen, Wei-Ming and Wang, Wei-Chen and Xiao, Guangxuan and Dang, Xingyu and Gan, Chuang and Han, Song},
 booktitle = {Proceedings of Machine Learning and Systems},
 editor = {P. Gibbons and G. Pekhimenko and C. De Sa},
 pages = {87--100},
 title = {AWQ: Activation-aware Weight Quantization for On-Device LLM Compression and Acceleration},
 url = {https://proceedings.mlsys.org/paper_files/paper/2024/file/42a452cbafa9dd64e9ba4aa95cc1ef21-Paper-Conference.pdf},
 volume = {6},
 year = {2024}
}

@inproceedings{ashkboos2024quarot,
title={QuaRot: Outlier-Free 4-Bit Inference in Rotated {LLM}s},
author={Saleh Ashkboos and Amirkeivan Mohtashami and Maximilian L. Croci and Bo Li and Pashmina Cameron and Martin Jaggi and Dan Alistarh and Torsten Hoefler and James Hensman},
booktitle={The Thirty-eighth Annual Conference on Neural Information Processing Systems},
year={2024},
url={https://openreview.net/forum?id=dfqsW38v1X}
}

@inproceedings{hu2022lora,
  author    = {Edward J. Hu and Yelong Shen and Phillip Wallis and Zeyuan Allen-Zhu and Yuanzhi Li and Shean Wang and Weizhu Chen},
  title     = {LoRA: Low-Rank Adaptation of Large Language Models},
  booktitle = {Proceedings of the 10th International Conference on Learning Representations (ICLR)},
  year      = {2022}
}

@inproceedings{liao-etal-2024-apiq,
    title = "{A}pi{Q}: Finetuning of 2-Bit Quantized Large Language Model",
    author = "Liao, Baohao  and
      Herold, Christian  and
      Khadivi, Shahram  and
      Monz, Christof",
    editor = "Al-Onaizan, Yaser  and
      Bansal, Mohit  and
      Chen, Yun-Nung",
    booktitle = "Proceedings of the 2024 Conference on Empirical Methods in Natural Language Processing",
    month = nov,
    year = "2024",
    address = "Miami, Florida, USA",
    publisher = "Association for Computational Linguistics",
    url = "https://aclanthology.org/2024.emnlp-main.1168/",
    doi = "10.18653/v1/2024.emnlp-main.1168",
    pages = "20996--21020",
}

@article{lecun1998mnist,
  title={The MNIST database of handwritten digits},
  author={LeCun, Yann and Cortes, Corinna},
  journal={Available: http://yann. lecun. com/exdb/mnist/},
  volume={24},
  year={1998}
}

\newpage
\newpage

\appendix

\section*{\huge{Supplementary Materials}}

\setcounter{section}{0}
\renewcommand{\thesection}{\Alph{section}}
\section{Additional Hessian Visualizations}
\label{sec:supplementary_hessians}

\begin{figure*}[t]
    \centering
    \includegraphics[width=\linewidth]{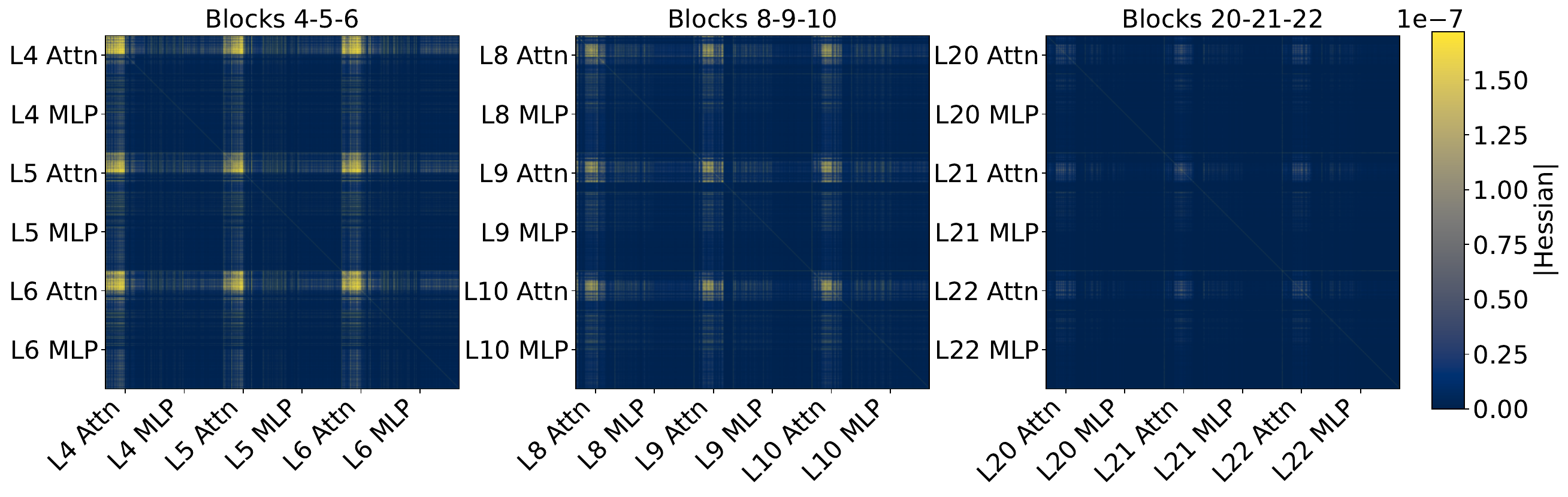}
    \caption{Our Kronecker--Fisher approximations for groups of Transformer
    blocks 4--5--6 (left), 8--9--10 (middle), and 20--21--22 (right) of
    OPT-350M. Yellow denotes high Hessian values and blue denotes low values.
    While the Hessian values are high in the earlier Transformer blocks, they
    become significantly smaller in the later blocks.}
    \label{fig:supplementary_3_hess}
\end{figure*}

Figure~\ref{fig:supplementary_3_hess} shows how the Hessian structure changes
with depth. The earlier and middle groups contain more pronounced high-value
regions and inter-layer correlations, whereas these values are substantially
weaker in the final group of Transformer blocks.

\begin{figure*}[t]
    \centering
    \includegraphics[width=\linewidth]{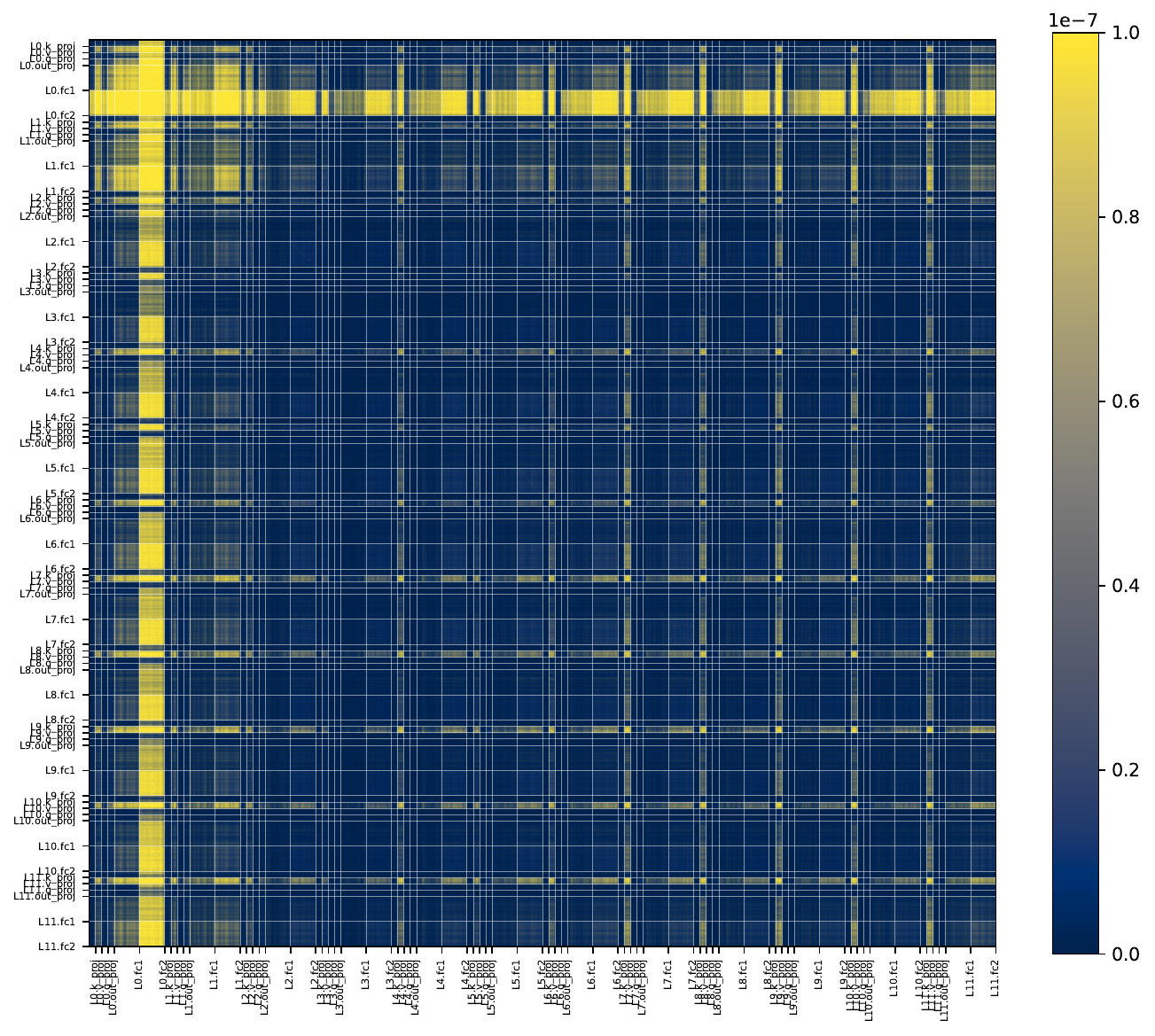}
    \caption{Full-model Kronecker--Fisher Hessian approximation for OPT-125M.
    This is one of the few language-model Hessians that can be presented in
    full while keeping the individual Transformer layers distinguishable.}
    \label{fig:supplementary_opt125m_full_hessian}
\end{figure*}

Figure~\ref{fig:supplementary_opt125m_full_hessian} provides a global view of
the Hessian across all OPT-125M Transformer blocks. Because OPT-125M is small
enough for the complete approximation to remain legible, the figure exposes
both the repeated block structure along the diagonal and the correlations
between layers in different blocks. For larger models, a full-model plot at
the same scale would make the layer boundaries indistinguishable, which is why
the main paper presents block-level and multi-block views instead.

\section{Pairwise Quantization with Concrete Values}
\label{sec:supplementary_pairwise_quantization}

\begin{figure*}[t]
    \centering
    \includegraphics[width=\linewidth]{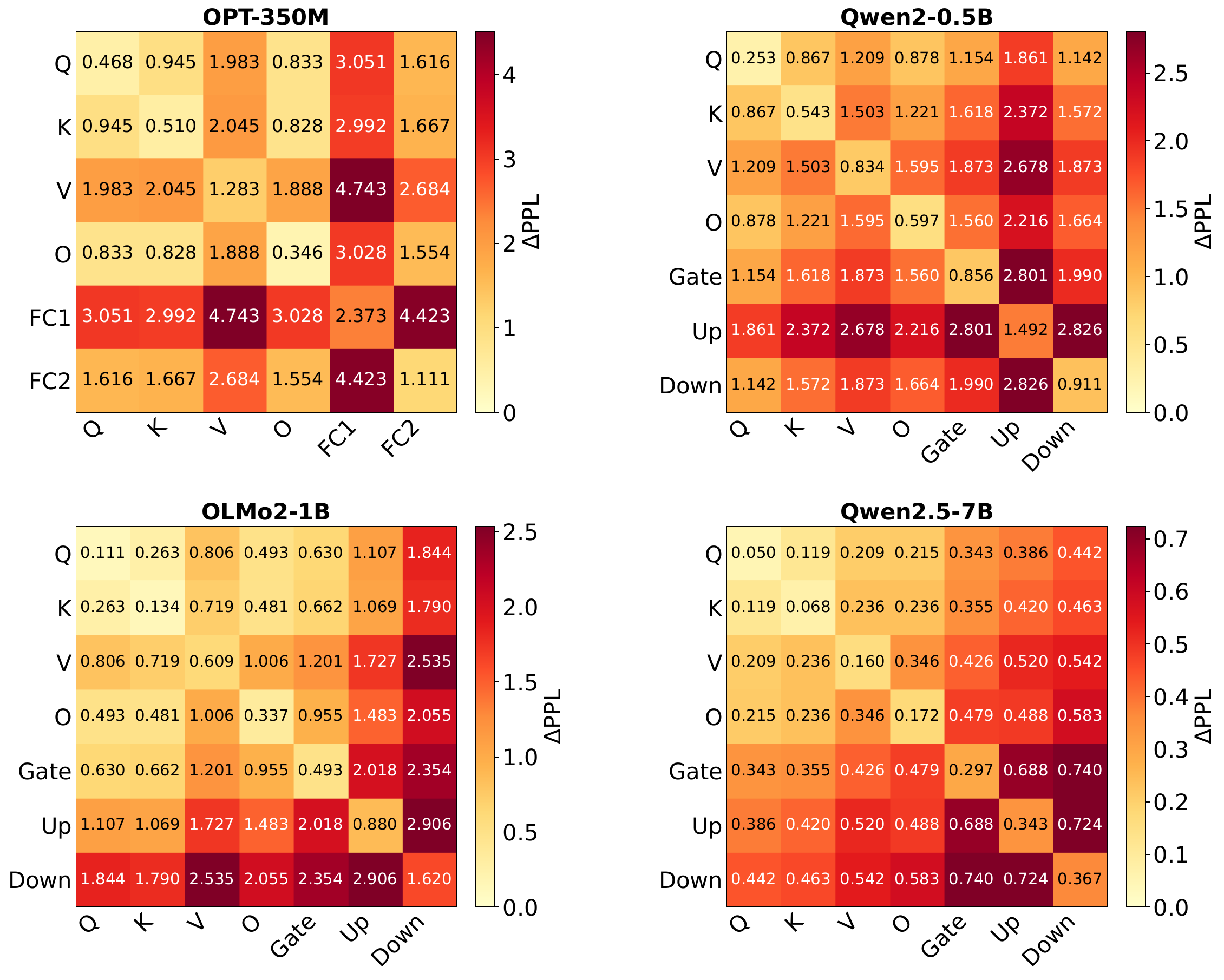}
    \caption{Pairwise 4-bit quantization results with the concrete perplexity
    increases shown in every cell. This is the annotated version of
    Figure~4 in the main paper. Each cell at row $i$ and
    column $j$ reports the perplexity increase after jointly quantizing layer
    types $i$ and $j$ in the middle Transformer blocks.}
    \label{fig:supplementary_pairwise_quant_values}
\end{figure*}

Figure~\ref{fig:supplementary_pairwise_quant_values} contains the same
pairwise-quantization experiment as Figure~4, but places the
measured values directly into the heatmaps. The annotations make it possible
to compare particular layer pairs quantitatively, while the color scale makes
the overall interaction pattern easier to see. As in the main-paper figure,
pairs involving the $V$-projection and pairs combining the upscale and
downscale layers produce some of the largest perplexity increases.
\section{Additional Experimental Results}
\subsection{Zero-Shot Results on Downstream Tasks}
\label{sec:supplementary_zero_shot}

We additionally evaluate whether the layer sensitivities observed through
WikiText-2 perplexity transfer to downstream tasks. For each model, we apply
$90\%$ sparsification to one layer type at a time in the middle
Transformer blocks, following the setup in the main paper, and evaluate the
resulting model without task-specific fine-tuning on PIQA, WinoGrande,
HellaSwag, ARC-Easy, and ARC-Challenge. Table~\ref{tab:zero_shot_sparsification}
reports the \textbf{accuracy decrease} relative to the dense baseline in percentage
points; thus, larger positive values indicate worse performance and greater sensitivity. The
baseline rows report absolute accuracy, also in percent, and the Avg. column is
the unweighted mean over the five tasks.

\begin{table*}[t]
    \centering
    \caption{Zero-shot accuracy under layer-type-wise $90\%$ sparsification.
    The baseline rows give dense-model accuracy (\%), while all projection rows
    give the \textbf{accuracy decrease} from that baseline in percentage points. Larger values indicate worse performance. Bold values mark
    the largest average decrease for each model.}
    \label{tab:zero_shot_sparsification}
    \begin{tabular}{llrrrrrr}
        \toprule
        Model & Setting & PIQA & WinoGrande & HellaSwag & ARC-E & ARC-C & Avg. \\
        \midrule
        OPT-350M & Baseline & 64.25 & 52.33 & 36.80 & 40.11 & 23.72 & 43.44 \\
          & $Q$   & 4.68 & 2.13 & 3.33 & 8.21 & -1.45 & 3.38 \\
          & $K$   & 2.34 & -1.03 & 3.51 & 6.86 & -0.09 & 2.32 \\
          & $V$   & 0.60 & 1.34 & 2.88 & 1.14 & 0.09 & 1.21 \\
          & $O$   & 0.05 & 0.95 & 2.68 & 1.73 & 0.17 & 1.12 \\
          & FC1   & 10.72 & 1.82 & 10.38 & 10.52 & 0.51 & \textbf{6.79} \\
          & FC2   & 3.59 & -0.08 & 3.25 & 4.17 & 0.43 & 2.27 \\
        \midrule
        Qwen2-0.5B & Baseline & 69.37 & 57.54 & 49.10 & 50.42 & 28.75 & 51.04 \\
          & $Q$    & 3.05 & 4.66 & 7.68 & 4.17 & 4.01 & 4.71 \\
          & $K$    & 4.08 & 4.58 & 9.26 & 7.03 & 3.24 & 5.64 \\
          & $V$    & 11.15 & 7.81 & 17.90 & 14.44 & 7.25 & \textbf{11.71} \\
          & $O$    & 3.21 & 4.42 & 7.28 & 4.29 & 3.67 & 4.57 \\
          & Gate   & 2.77 & 4.89 & 7.72 & 2.78 & 1.54 & 3.94 \\
          & Up     & 7.24 & 5.13 & 14.00 & 9.97 & 5.46 & 8.36 \\
          & Down   & 6.31 & 3.63 & 9.60 & 7.95 & 4.52 & 6.40 \\
        \midrule
        OLMo2-1B & Baseline & 74.81 & 63.85 & 66.61 & 72.60 & 40.96 & 63.77 \\
          & $Q$    & 2.67 & 4.81 & 10.11 & 11.83 & 7.34 & 7.35 \\
          & $K$    & 2.12 & 3.95 & 4.85 & 9.47 & 6.57 & 5.39 \\
          & $V$    & 2.01 & 2.29 & 8.32 & 8.71 & 5.38 & 5.34 \\
          & $O$    & 3.10 & 4.50 & 13.14 & 15.28 & 9.30 & 9.06 \\
          & Gate   & 2.77 & 5.45 & 10.11 & 13.17 & 6.14 & 7.53 \\
          & Up     & 8.76 & 10.89 & 22.11 & 26.89 & 12.37 & \textbf{16.20} \\
          & Down   & 7.73 & 4.97 & 19.20 & 25.17 & 13.05 & 14.02 \\
        \midrule
        Qwen2.5-7B & Baseline & 79.38 & 70.56 & 78.21 & 75.93 & 49.66 & 70.75 \\
          & $Q$    & 1.80 & 4.89 & 3.87 & 1.26 & 1.79 & 2.72 \\
          & $K$    & 1.09 & 4.81 & 6.84 & 7.37 & 3.16 & 4.65 \\
          & $V$    & 1.09 & 1.82 & 5.93 & 3.32 & 3.75 & 3.18 \\
          & $O$    & 1.25 & 4.66 & 7.71 & 5.85 & 4.01 & 4.70 \\
          & Gate   & 25.35 & 21.70 & 49.16 & 47.01 & 27.39 & \textbf{34.12} \\
          & Up     & 5.66 & 7.81 & 15.26 & 7.03 & 8.11 & 8.77 \\
          & Down   & 3.92 & 5.13 & 10.71 & 5.43 & 7.42 & 6.52 \\
        \bottomrule
    \end{tabular}
\end{table*}

The downstream results recover the principal sensitivity patterns found with
perplexity. FC1 is the most damaging projection to sparsify in OPT-350M; the
$V$-projection is the most damaging in Qwen2-0.5B; the upscale and downscale
projections dominate in OLMo2-1B; and the gate projection is by far the most
sensitive component in Qwen2.5-7B. The precise magnitude varies by task---in
particular, HellaSwag and ARC-Easy often show the largest drops---but the most
sensitive layer types remain consistent across evaluation criteria. Negative
values for a few OPT-350M task--projection pairs denote small measured
improvements and do not change the aggregate ranking.


\end{document}